\documentclass[journal]{IEEEtran}

\usepackage{cite}

\ifCLASSINFOpdf
\else
\fi
\usepackage{amsmath}
\usepackage{array}

\ifCLASSOPTIONcompsoc
  \usepackage[caption=false,font=normalsize,labelfont=sf,textfont=sf]{subfig}
\else
  \usepackage[caption=false,font=footnotesize]{subfig}
\fi
\usepackage{fixltx2e}
\usepackage{amsmath,amssymb,amsfonts,stmaryrd}
\usepackage[latin1]{inputenc}
\usepackage{hyperref}
\usepackage{graphicx}
\usepackage{color}

\newcommand{\esp}{\mathrm{E}}

\newcommand{\matr}[1]{\boldsymbol{#1}}
\newcommand{\pr}{\mathrm{p}}
\newcommand{\tr}{\mathrm{tr}}
\newcommand{\ud}{\mathrm{d}}
\newcommand{\vect}[1]{\boldsymbol{#1}}

\newcommand{\gcite}{\cite}
\newcommand{\gcitet}{\cite}

\newcommand{\syst}[1]{\mathcal{#1}}
\newcommand{\depXYD}[1]{\mathfrak{B}_{\mathrm{#1}} ( \syst{X}, \syst{Y} | D )}

\newcommand{\DKL}{D_{\mathrm{KL}}}

\newcommand{\hi}{^{(0)}}
\newcommand{\hd}{^{(1)}}
\newcommand{\hdi}{^{(i)}}
\newcommand{\paramhi}{\vect{\theta}\hi}
\newcommand{\paramhd}{\vect{\theta}\hd}
\newcommand{\paramhdi}{\vect{\theta}\hdi}
\newcommand{\estparamhi}{\hat{\vect{\theta}}_N\hi}
\newcommand{\estparamhd}{\hat{\vect{\theta}}_N\hd}
\newcommand{\estparamhdi}{\hat{\vect{\theta}}_N\hdi}
\newcommand{\limparamhi}{\vect{\theta}_{\infty}\hi}
\newcommand{\limparamhd}{\vect{\theta}_{\infty}\hd}

\newcommand{\modif}[1]{#1}

\newcommand{\supplmat}[1]{#1 of Supplementary Material}

\graphicspath{{Figures/}}

\begin{document}
%
\title{An inferential measure of dependence between two systems using Bayesian model comparison}
%
%
%

\author{Guillaume Marrelec
        and~Alain Giron
\thanks{G. Marrelec and A. Giron are with the Laboratoire d'imagerie biom{\'e}dicale, LIB, Sorbonne Université, CNRS, INSERM, F-75006, Paris, France. email: firstname.lastname@inserm.fr.}
}

\maketitle

\begin{abstract}
 \modif{We propose to quantify dependence between two systems} $\syst{X}$ and $\syst{Y}$ in a dataset $D$ based on the Bayesian comparison of two models: one, $H_0$, of statistical independence and another one, $H_1$, of dependence. In this framework, \modif{dependence} between $\syst{X}$ and $\syst{Y}$ in $D$, denoted $\depXYD{}$, is quantified as $P ( H_1 | D )$, the posterior probability for the model of dependence given $D$, or any strictly increasing function thereof. It is therefore a measure of the \modif{evidence for} dependence between $\syst{X}$ and $\syst{Y}$ as modeled by $H_1$ and observed in $D$. We review several statistical models and reconsider standard results in the light of $\depXYD{}$ as a \modif{measure of dependence}. Using simulations, we focus on two specific issues: the effect of noise and the behavior of $\depXYD{}$ when $H_1$ has  a parameter coding for the intensity of dependence. We then derive some general properties of $\depXYD{}$, showing that \modif{it} quantifies the information contained in $D$ in favor of $H_1$ versus $H_0$. While some of these properties are typical of what is expected from a valid measure of dependence, others are novel and naturally appear as desired features for specific measures of dependence, which we call \emph{inferential}. We finally put these results in perspective; in particular, we discuss the consequences of using the Bayesian framework as well as the similarities and differences between $\depXYD{}$ and mutual information.
\end{abstract}
 

%
\IEEEpeerreviewmaketitle

\section{Introduction}
%
%
%
%

\IEEEPARstart{I}{ndependence} and dependence are key concepts in science whose goal is to characterize the structural relationships between systems. Consider two systems $\syst{X}$ and $\syst{Y}$ characterized by a probabilistic description in terms of (possibly multivariate) random variables $X$ and $Y$, respectively, with known joint probability distribution $f_{ X Y } ( \vect{x}, \vect{y} )$ and marginals $f_X ( \vect{x} )$ and $f_Y ( \vect{y} )$. In this context, $\syst{X}$ and $\syst{Y}$ are said to be independent if they underlying random variables are independent, i.e., \gcite[\S~2.2]{Whittaker-1990}
\begin{equation} \label{eq:indep:def}
 f_{XY} ( \vect{x}, \vect{y} ) = f_X ( \vect{x} ) \, f_Y ( \vect{y} ).
\end{equation}
When $X$ and $Y$ are not independent, $f_{XY} ( \vect{x}, \vect{y} )$ differ from $f_X ( \vect{x} ) \, f_Y ( \vect{y} )$, and both systems are said to be dependent. In this case, an important issue is the \modif{quantification of dependence} between $\syst{X}$ and $\syst{Y}$, where one tries to \modif{measure to what degree $f_{ X Y } ( \vect{x}, \vect{y} )$ differs} from $f_X ( \vect{x} ) \, f_Y ( \vect{y} )$. \modif{There are many ways to depart from independence, and this question remains open in the general case. Many} measures of dependence have been proposed, including, but not limited to, mutual information \gcite[\S2 and \S8]{Cover_TM-1991}, maximal correlation coefficient \gcite{Gebelein-1941, Renyi-1959b}, the mixed derivative measure of marginal interaction \gcite[\S2.3]{Whittaker-1990}, Hoeffding's procedure \gcite{Hoeffding-1948}, distance correlation \gcite{Szekely-2007, Edelmann-2021}, circular correlation \gcite{Jupp-1980}, Hilbert--Schmidt information criterion \gcite{Gretton-2005b}.
\par
Once a theoretical measure has been proposed, another layer of complexity is often added by the fact that we do not know $f_{ X Y } ( \vect{x}, \vect{y} )$. A common situation is when one knows (or assumes) that it belongs to a family $f_{ X Y } ( \vect{x}, \vect{y} | \vect{\theta} )$ parameterized by an (unknown) parameter $\vect{\theta}$ which has to be estimated using a dataset of $N$ independent and identically distributed (i.i.d.) samples $( \vect{x}_n, \vect{y}_n )$. Various estimation strategies have been proposed, from plug-in estimators (where one computes the measure by replacing the true value of the parameter by its estimator) \gcite[Chap.~9]{Anderson_TW-1958}, \gcite[Chap.~12, \S3.6]{Kullback-1968} to more refined techniques proposing estimators for the measure itself \gcite{Paninski-2003, Kraskov-2004, Kraskov-2011}. The interest of this kind of approaches relies on the asymptotic convergence of the estimators towards the true value of the measure. However, the values taken by these estimators for finite data size do not have simple interpretations.
\par
In the present paper, we introduce another general measure of dependence $\depXYD{}$ between systems $\syst{X}$ and $\syst{Y}$ given dataset $D$. Our starting point is the real-life situation where the joint behavior of $( \syst{X}, \syst{Y} )$ is characterized by a dataset of $N$ i.i.d. samples $( \vect{x}_n, \vect{y}_n )$. We propose to characterize \modif{dependence} between $\syst{X}$ and $\syst{Y}$ in $D$ by using a Bayesian inference scheme that compares the credibility of two competing models: $H_0$, which describes $\syst{X}$ and $\syst{Y}$ as independent, and $H_1$, which describes them as dependent. This setting was already used to infer the structure of independence within a multivariate distribution \gcite{Marrelec-2015, Marrelec-2021b}. In the case of bivariate discrete distributions and multivariate normal distributions, a connection was also observed between the log posterior odd ratio and mutual information \gcite{Wolf_DR-1994, Marrelec-2015, Marrelec-2021b}. We here propose to go one step further and advocate that the  Bayesian comparison of models $H_1$ and $H_0$ mentioned above provides a family of measures that can be used to quantify the level of dependence between $\syst{X}$ and $\syst{Y}$. As will be further developed below, any measure $\depXYD{}$ in this family is an increasing function of correlation, mutual information, the minimum discrimination information, and the log-likelihood ratio criterion for testing independence, in cases where it makes sense to define such measures. As a consequence, $\depXYD{}$ shares many features with these classical measures of dependence. Unlike these same measures, though, the value of $\depXYD{}$ for a given $D$ and finite $N$ has a direct, exact interpretation as (a strictly increasing function of) the probability that $X$ and $Y$ be dependent (as described by $H_1$) for $D$. In other words, $\depXYD{}$ quantifies the evidence for---or credibility of---dependence between $\syst{X}$ and $\syst{Y}$ as modeled by $H_1$ and observed in $D$, i.e., the amount of information contained in $D$ in favor of $H_1$ versus $H_0$. It has several interesting features regarding the effect of data size and noise, which are detailed throughout the manuscript and characterize what we would call an \emph{inferential} measure of dependence.
\par
The outline of our manuscript is the following. In Section~\ref{s:bmod}, we present the general theoretical framework. In Section~\ref{s:ss}, we use simulation studies to investigate two specific issues: the effect of noise and the behavior of $\depXYD{}$ when $H_1$ has  a parameter coding for the intensity of dependence. Section~\ref{s:rla} provides a real-life application in neuroscience and neuroimaging. We then state some key properties of $\depXYD{}$ in Section~\ref{s:gp}. Further issues are discussed in Section~\ref{s:disc}.


\section{Bayesian measures of dependence} \label{s:bmod}

In this section, we present the core of our method. After a quick review of Bayesian model comparison for the investigation of statistical independence and dependence (Section~\ref{ss:bmod:mc}), we introduce a general measure of dependence $\depXYD{}$ which can either take the form of a posterior probability or any strictly increasing function thereof (Section~\ref{ss:bmod:afom}). We then investigate the theoretical properties of $\depXYD{lnr}$, a particular instance of $\depXYD{}$, in the case of known distributions (Section~\ref{ss:bmod:kd}) and known likelihood functions with unknown parameters (Section~\ref{ss:bmod:klfwup}). In Section~\ref{ss:bmod:nm}, we apply our framework to the important case where $H_0$ is nested in $H_1$, a framework that can be applied to several common models, such as maximum-entropy distributions, multivariate normal distributions, and bivariate discrete distributions. In Section~\ref{ss:bmod:cm}, we consider the other particular case where dependence is modeled through a copula. Section~\ref{ss:bmod:mm} investigates the consequences of model misspecification. Finally, in Section~\ref{ss:s}, we  summarize the main results obtained thus far.

\subsection{Model comparison} \label{ss:bmod:mc}

We here provide a quick description of the general framework of Bayesian model comparison for the investigation of statistical dependence. For more details, the reader can refer to \gcitet{Marrelec-2015} or \gcitet{Marrelec-2021b}. Consider the following two competing models:
\begin{itemize}
 \item A model $H_0$ (with parameter $\paramhi \in \Theta\hi$) in which $\syst{X}$ and $\syst{Y}$ are independent and where the likelihood is given by
 \begin{equation}
   l_0 ( \vect{x}, \vect{y} | \paramhi ) = f_X\hi ( \vect{x} | \paramhi ) \, f_Y\hi ( \vect{y} | \paramhi );
 \end{equation}
 \item Another model $H_1$ (with model parameter $\paramhd \in \Theta\hd$) in which $\syst{X}$ and $\syst{Y}$ are dependent and the likelihood is
 \begin{equation}
   l_1 ( \vect{x}, \vect{y} | \paramhd ) = f_{ X Y }\hd ( \vect{x}, \vect{y} | \paramhd ).   
 \end{equation}
\end{itemize}
In a Bayesian framework, all the information from the data that is relevant for the problem at hand is summarized  by the posterior probabilities of $H_0$ and $H_1$ given the dataset $D$, $\pr ( H_0 | D )$ and $\pr ( H_1 | D )$, respectively. Both quantities can be calculated using Bayes updating rule
\begin{equation} \label{eq:bayes}
 \pr ( H_i | D ) = \frac{ \pr ( H_i ) \, \pr ( D | H_i ) } { \pr ( D ) }, \qquad i = 0, 1.
\end{equation}
$\pr ( H_i )$ is the model prior probability and can be set depending on the prior belief that we have regarding the relative plausibility of both competing hypotheses. As to the marginal model likelihood $\pr ( D | H_i )$, it can be expressed using the marginalization formula
\begin{equation} \label{eq:vraismarg}
 \pr ( D | H_i ) = \int_{ \paramhdi \in \Theta\hdi } \pr ( \paramhdi | H_i ) \, \pr ( D | H_i, \paramhdi ) \,  \ud \paramhdi.
\end{equation}
In this expression, $\pr ( \paramhdi | H_i )$ is the parameter prior. We set it to $h_i ( \paramhdi )$ for $\paramhdi \in \Theta\hdi$. $\pr ( D | H_i, \paramhdi )$ is the likelihood function. For $N$ i.i.d. samples,
\begin{equation}
  D = \{ ( \vect{x}_1, \vect{y}_1 ), \dots, ( \vect{x}_N, \vect{y}_N ) \},
\end{equation}
it can be decomposed as
\begin{equation} \label{eq:vrais}
 \pr ( D | H_i, \paramhdi ) = \prod_{ n = 1 } ^ N l_i ( \vect{x}_n, \vect{y}_n | \paramhdi ).
\end{equation}

\subsection{A family of measures} \label{ss:bmod:afom}

By definition, $\pr ( H_1 | D )$ quantifies the posterior probability that $\syst{X}$ and $\syst{Y}$ be dependent. As a consequence, it can be considered as a measure of dependence between $\syst{X}$ and $\syst{Y}$,
\begin{equation} \label{eq:def:prob}
 \depXYD{pr} \equiv \pr ( H_1 | D ).
\end{equation}
Indeed, it is equal to 0 when it is known that $\syst{X}$ and $\syst{Y}$ are independent as in $H_0$, equal to 1 when they are known to be dependent in agreement with $H_1$, and in between when we are sure of neither; the larger it is, the more probable the dependence of $\syst{X}$ and $\syst{Y}$ is.
\par
If one agrees to treat $\pr ( H_1 | D )$ as a measure of dependence, then various strictly increasing mappings of it could also be considered, with different ranges. For instance, one could use the posterior odd ratio of $H_1$ versus $H_0$
\begin{equation} \label{eq:def:r}
 \depXYD{r} \equiv \frac{ \pr ( H_1 | D ) } { \pr ( H_0 | D ) } = \frac{ \pr ( H_1 ) } { \pr ( H_0 ) } \, \frac{ \pr ( D | H_1 ) } { \pr ( D | H_0 ) }.
\end{equation}
The first fraction of the right-hand side equation is the so-called prior odd ratio, while the second fraction is the Bayes factor. The Bayes factor itself could be used as a measure of dependence. We can also use the log scale to obtain a better representation of the measure of dependence, either in its usual form
\begin{equation} \label{eq:def:b}
 \depXYD{lnr} \equiv \ln \frac{ \pr ( H_1 ) } { \pr ( H_0 ) } + \ln \frac{ \pr ( D | H_1 ) } { \pr ( D | H_0 ) },
\end{equation}
or in \modif{$\log_{10}$,
so} that a value of $b$ means that $H_1$ is $10^b$ times more probable than $H_0$; or expressed such as to yield values on a scale similar to decibels \gcite{Good-1979}. Measures similar to $\depXYD{lnr}$, $\depXYD{logr}$ and $\depXYD{dB}$ but relying on $\depXYD{BF}$ instead of $\depXYD{r}$ could be proposed (i.e., not taking the relative priors of $H_0$ and $H_1$ into account).
\par
Importantly, all these measures are strictly increasing function of dependence, i.e., they increase as the probability for $H_1$ (dependence) increases.
\par
In the following, we will mostly focus on $\depXYD{lnr}$ and $\depXYD{logr}$, as it is these measures whose properties are easiest to investigate. Furthermore, they show the closest connections to existing frameworks for quantifying dependence, such as the log-likelihood criterion for independence and mutual information.

\subsection{Known distributions} \label{ss:bmod:kd}

If we assume that the distribution of $( X, Y )$ is known exactly in both $H_0$ and $H_1$, then it is straightforward to show that (see \supplmat{\S1}):
\begin{itemize}
 \item If $H_0$ is true (independence), $\depXYD{lnr}$ tends to $-\infty$ as $N \to \infty$, and its sampling expectation is a decreasing function of $N$.
 \item If $H_1$ is true (dependence), $\depXYD{lnr}$ tends to $+\infty$ as $N \to \infty$, and its expectation is an increasing function of $N$ and $I ( X, Y )$, the mutual information between $X$  and $Y$.
\end{itemize}
The rest of the section is devoted to show that these results also hold for more general models, albeit in a weaker, asymptotic form.

\subsection{Known likelihood functions with unknown parameters} \label{ss:bmod:klfwup}

We now consider the more general case where each likelihood function is not exactly known but belongs to a known family with unknown parameter. While an exact expression for $\depXYD{lnr}$ cannot be obtained in this case, we can still derive an asymptotic approximation in a fashion similar to \gcitet[\S\S7.22--7.27]{OHagan-2004}, which itself relies on the consistency of the maximum-likelihood estimate.
\par
For $i \in \{ 0, 1  \}$, let $L_i ( \paramhdi ) = \pr ( D | H_i, \paramhdi )$ be the likelihood function associated with model $H_i$, as defined in \eqref{eq:vrais}. The marginal model likelihoods of \eqref{eq:vraismarg} can formally be expressed as
\begin{equation} \label{eq:bmod:vraismarg}
  P ( D | H_i ) = \int_{ \paramhdi \in \Theta\hdi } h_i ( \paramhdi ) \, L_i ( \paramhdi ) \, \ud \paramhdi.
\end{equation}
We assume that $L_i ( \paramhdi )$ is unimodal and set $\estparamhdi$ its maximum-likelihood estimate,
\begin{equation} \label{eq:mv}
 \estparamhdi = \mathrm{argmax}_{ \paramhdi \in \Theta\hdi } L_i ( \paramhdi ), \qquad i = 0, 1.
\end{equation}
Assuming that the prior $h_i ( \paramhdi )$ is strictly positive and of slow variation around $\estparamhdi$, the integral of \eqref{eq:bmod:vraismarg} can be approximated using Laplace method \gcite{Tierney-1986, Gelfand-1994} (see \supplmat{\S2}), yielding
\begin{equation} \label{eq:bmod:klfwup:dep}
 \depXYD{lnr} = \ln \frac{ L_1 ( \estparamhd ) } { L_0 ( \estparamhi ) } - \frac{ D_2 - D_1 } { 2 } \ln N + O ( 1 ).
\end{equation}
and
\begin{eqnarray} \label{eq:bmd:klfwup:dep:3}
 \depXYD{lnr} & = & N \Bigg[ \hat{I} ( X, Y ) + \frac{ 1 } { N } \sum_{ n = 1 } ^ N \ln \frac{ f_X\hd ( \vect{x}_n | \estparamhd ) } { f_X\hi ( \vect{x}_n | \estparamhi )  } \nonumber \\
 & & \hspace{-3cm} + \frac{ 1 } { N } \sum_{ n = 1 } ^ N \ln \frac{ f_Y\hd ( \vect{y}_n | \estparamhd ) } { f_Y\hi ( \vect{y}_n | \estparamhi ) } \Bigg] - \frac{ D_2 - D_1 } { 2 } \ln N + O ( 1 ).
\end{eqnarray}
In \eqref{eq:bmod:klfwup:dep}, the first term of the right-hand side is the classical likelihood ratio test statistic, showing the connection between our approach and the log-likelihood ratio criterion for testing independence \gcite[\S9.2]{Anderson_TW-1958}. The second term of the right-hand side is the BIC correction \gcite{Schwarz-1978}. In \eqref{eq:bmd:klfwup:dep:3}, $\hat{I} ( X, Y )$ is the sampling mutual information  under $H_1$.
\par
In the absence of further assumption, nothing can be said about the asymptotic behavior of this quantity, which depends on the true likelihood function as well as the limits of the maximum-likelihood estimates $\estparamhi$ and $\estparamhd$. These limits, in turn, are not necessary well defined in the general case.

\subsection{Nested models} \label{ss:bmod:nm}

We here consider the particular case where $H_0$ is nested in $H_1$. In this case, the models allowed by $H_0$ are included in those allowed by $H_1$, ie., models of $H_0$ are particular cases of models in $H_1$. In other words, there exists a function $\pi$ such that, for any $\paramhi \in \Theta\hi$, we have
\begin{equation}
 l_0 ( \vect{x}, \vect{y} | \paramhi ) = l_1 [ \vect{x}, \vect{y} | \pi ( \paramhi ) ].
\end{equation}
For the sake of simplicity, it is often assumed that $\pi ( \paramhi )$ is a projection, i.e., $\Theta\hd$ can be parameterized by $\paramhd = ( \paramhi, \vect{\phi} )$ with $\paramhi \in \Theta\hi$ such that $\pi ( \paramhi ) = ( \paramhi, \vect{\phi}_0 )$ (i.e., $\vect{\phi} = \vect{\phi}_0$). Then it can be shown the following results (see \supplmat{\S3}).

\paragraph{Under $H_0$}

When $N \to \infty$, $\depXYD{lnr}$ essentially behaves as $- \frac{ 1 } { 2 } ( D_1 - D_0 ) \ln N$, which is a decreasing function of $N$ that tends to $-\infty$. Also, $\esp [ \depXYD{lnr} | H_ 0 ]$ is a decreasing function of $N$.

\paragraph{Under $H_1$}

When $N \to \infty$, $\depXYD{lnr}$ is approximately  linearly increasing in $N$. $\esp [ \depXYD{lnr} | H_ 1 ]$ has a first-order approximation that is an increasing function of $N$, but a second-order approximation that may first decrease before it increases.

\par
Nested models in the particular cases of maximum-entropy distributions, multivariate normal distributions, and bivariate discrete distributions are considered in \supplmat{\S4, \S5, and \S6}, respectively, together with details about the connection between our method and existing methods, such as the log-likelihood ratio criterion, the BIC, mutual information, and the minimum discrimination information statistic.

\subsection{Copula models} \label{ss:bmod:cm}

Another particular case of interest is when $H_0$ and $H_1$ share the same assumptions regarding marginals for $X$ and $Y$---modeled by $f_X ( \vect{x} | \vect{\phi} )$ and $f_Y ( \vect{y} | \vect{\phi} )$, respectively---and $H_1$ models dependence through a copula with density $c ( \vect{u}, \vect{v} | \vect{\psi})$ \gcite[Chap.~8]{Ruppert-2015}. The model parameters are therefore $\paramhi = \vect{\phi}^{(0)}$ for $H_0$ and $\paramhd = ( \vect{\phi}^{(1)}, \vect{\psi}^{(1)} )$ for $H_1$. A common approach to estimate parameters is the method of inference function for margins (IFM) \gcite[\S10.1]{Joe-1997}, which has a simple interpretation in our framework: First estimate $\paramhi$ using the maximum-likelihood estimate $\estparamhi = \hat{\vect{\phi}}_N^{(0)}$, and then estimate $\paramhd$ by first setting $\hat{\vect{\phi}}_N^{(1)} = \hat{\vect{\phi}}_N^{(0)}$ and then finding the maximum-likelihood estimate $\estparamhd = ( \hat{\vect{\phi}}_N^{ ( 0 ) }, \hat{\vect{\psi}}_N^{ ( 1 ) } ) $. 
In this case, \eqref{eq:bmd:klfwup:dep:3} simplifies to
\begin{equation} \label{eq:bmod:cm:dep}
 \depXYD{lnr} = N \hat{I} ( X, Y ) - \frac{ D_2 - D_1 } { 2 } \ln N + O ( 1 ),
\end{equation}
with
\begin{equation} \label{eq:bmod:cm:im}
 \hat{I} ( X, Y ) = \frac{ 1 }{ N } \sum_{ n = 1 } ^ N \ln c \left[ F_X ( \vect{x}_n | \hat{\vect{\phi}}_N^{(0)} ), F_Y ( \vect{y}_n | \hat{\vect{\phi}}_N^{(0)} ) \Big| \hat{\vect{\psi}}_N^{ ( 1 ) } \right],
\end{equation}
where $F_X ( \vect{x} |\vect{\phi} )$ and $F_Y ( \vect{y} | \vect{\phi} )$ are the cumulative distribution functions of $x$ and $Y$, respectively. This quantity is related to minus the entropy of $c ( \vect{u}, \vect{v} | \vect{\psi})$ \gcite{Ma_J-2011}.

\subsection{Model misspecification} \label{ss:bmod:mm}

We now investigate the consequence of considering a true underlying generative model that is neither $H_0$ nor $H_1$. To this aim, we assume that the generative distribution is $f ( \vect{x}, \vect{y} | \vect{\theta} )$, that the estimators of the model parameters under $H_0$ and $H_1$ have limits, i.e., $\estparamhi \stackrel{N \to \infty}{\to} \limparamhi$ and $\estparamhd \stackrel{N \to \infty}{\to} \limparamhd$, and that these limits are such that Laplace approximation can be applied. In \eqref{eq:bmod:klfwup:dep}, the leading term of $\depXYD{lnr}$ is $\ln [ L_1 ( \estparamhd ) / L_0 ( \estparamhi ) ]$, which can be expressed as (see \supplmat{\S7})
\begin{eqnarray}
 & &  N \Big\{ \DKL \left[ f ( \vect{x}, \vect{y} | \vect{\theta} ) \| f_X\hi ( \vect{x} | \limparamhi ) \, f_Y\hi ( \vect{y} | \limparamhi ) \right] \nonumber \\
 & & \hspace{-.5cm} - \DKL \left[ f ( \vect{x}, \vect{y} | \vect{\theta} ) \| f_{XY}\hd ( \vect{x}, \vect{y} | \limparamhd ) \right] \Big\} + o ( N ),
\end{eqnarray}
where $o ( \cdot )$ is the usual little-o notation. If both Kullback--Leibler divergences in this equation differ, then the likelihood ratio is roughly linear in $N$, with a proportionality factor whose sign is given by the difference in Kullback--Leibler divergences. As a consequence, for $N$ large enough, if the model corresponding to $H_0$ is closer to the true generative model (as measured by Kullback--Leibler divergence), $\depXYD{lnr}$ will be decreasing and will tend to $-\infty$; if it is the model corresponding to $H_1$ that is closer, $\depXYD{lnr}$ will be increasing and will tend to $+\infty$.

\subsection{Summary} \label{ss:s}

In Section~\ref{s:bmod}, we introduced a general measure $\depXYD{}$ to quantify statistical dependence in data sets. Our framework is based on the Bayesian comparison of a model $H_1$ taking dependence into account (in a form specified by the model) and a model $H_0$  not taking this dependence into account (Sections \ref{ss:bmod:mc}). We then defined $\depXYD{}$ as $\pr ( H_1 | D )$ or any strictly increasing functions thereof (Section~\ref{ss:bmod:afom}). We investigated the behavior of these measures on i.i.d. data in the case of a known distribution (Section~\ref{ss:bmod:kd}) or a known likelihood family with unknown parameters (Section~\ref{ss:bmod:klfwup}). We then delved into two particular cases: nested models (Section~\ref{ss:bmod:nm})---which can be applied to maximum-entropy distributions, multivariate normal distributions, and bivariate discrete distributions---, and copula models of dependence (Section~\ref{ss:bmod:cm}). We finally considered the consequences of model misspecification (Section~\ref{ss:bmod:mm}). In all the cases, we showed that $\depXYD{lnr}$ asymptotically behaved as follows:
\begin{itemize}
 \item Under $H_0$, $\depXYD{lnr}$ is a decreasing function of $N$ which tends to $- \infty$ as $N \to \infty$;
 \item Under $H_1$, $\depXYD{lnr}$ is an increasing function of $N$ (possibly after an initial decrease) which tends to $+ \infty$ as $N \to \infty$.
 \item $\depXYD{lnr}$ is an increasing function of $\hat{I} ( X, Y )$ which tends to $+ \infty$ as $I ( X, Y ) \to + \infty$.
 \item In case of model misspecification, $\depXYD{lnr}$ behaves as if the model closer to the true generative one (in terms of Kullback--Leibler divergence) were the true one.
\end{itemize}
While these results involve developments that are standard in statistical theory and information theory, they shed some important light on the relevance of $\depXYD{}$ as a valid measure of \modif{dependence}. We will come back to this point in Section~\ref{s:gp}.

\section{Simulation study} \label{s:ss}

In the previous section, we provided general results regarding some common statistical models of dependence where we had direct access to the variables of interest. Here, we use synthetic data to focus on two specific issues: the effect of noise, and the behavior of $\depXYD{}$ when $H_1$ has  a parameter coding for the intensity of dependence.
\par
To investigate the effect of noise, we considered synthetic data originating from three distinct models: two variables following a bivariate normal distribution plus noise (Section~\ref{ss:ss:mbb}), two variables related by a functional relationship plus noise (Section~\ref{ss:ss:fd}), and two chaotic systems (Section~\ref{ss:ss:dotcs}). In all three examples, we varied the size of the dataset $N$ and the variance of the noise $\sigma^2$. We predicted that a good measure of \modif{dependence} should behave as follows: As the dataset becomes more and more informative (i.e., as $N$ increases and $\sigma^2$ decreases), the value of $\depXYD{}$ should (i) decrease and get increasingly closer to its lower bound if the true underlying model is a model of independence ($H_0$);  and (ii) increase and get increasingly closer to its upper bound if the true underlying model is a model of dependence ($H_1$).
\par
The behavior of $\depXYD{}$ when $H_1$ has  a parameter coding for the intensity of dependence was investigated using three models as well: the abovementioned model of bivariate normal distribution plus noise (Section~\ref{ss:ss:mbb}), a copula model of dependence (Section~\ref{ss:ss:dtc}), and the abovementioned model of two chaotic systems (Section~\ref{ss:ss:dotcs}). In the three models, the intensity of dependence under $H_1$ was quantified through a parameter ($\rho$ for Sections \ref{ss:ss:mbb} and \ref{ss:ss:dtc}, $C$ for Section~\ref{ss:ss:dotcs}). We varied the parameter and expected $\depXYD{logr}$ to be an increasing function of the intensity of dependence ($|\rho|$ for Sections \ref{ss:ss:mbb}) and \ref{ss:ss:dtc}, $C$ for Section~\ref{ss:ss:dotcs}).
\par
For all simulations, we focused on $\depXYD{logr}$, whose value is simple to interpret (a value of $b$ means that $H_1$ is $10^b$ times more probable than $H_0$). Its lower and upper bounds are $- \infty$ and $+ \infty$, respectively. A summary of the main results can be found in Section~\ref{ss:sor}.

\subsection{Bivariate normal distribution with noise} \label{ss:ss:mbb}

\begin{figure*}[!htbp]
 \centering
 \begin{tabular}{cc}
  \includegraphics[width=\columnwidth]{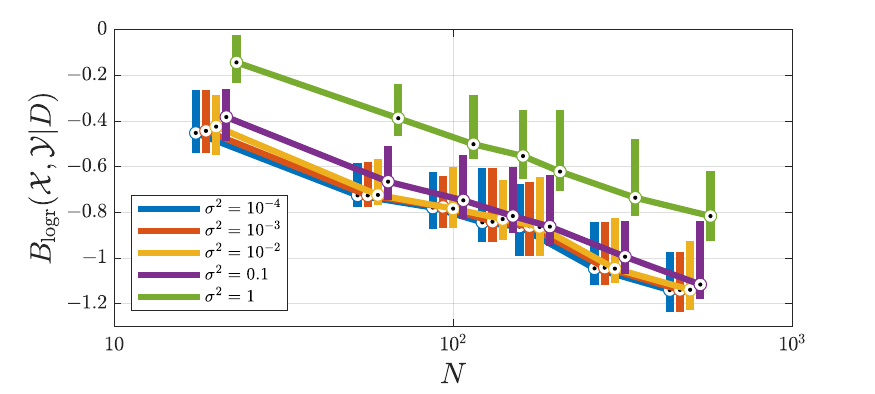}
  & \includegraphics[width=\columnwidth]{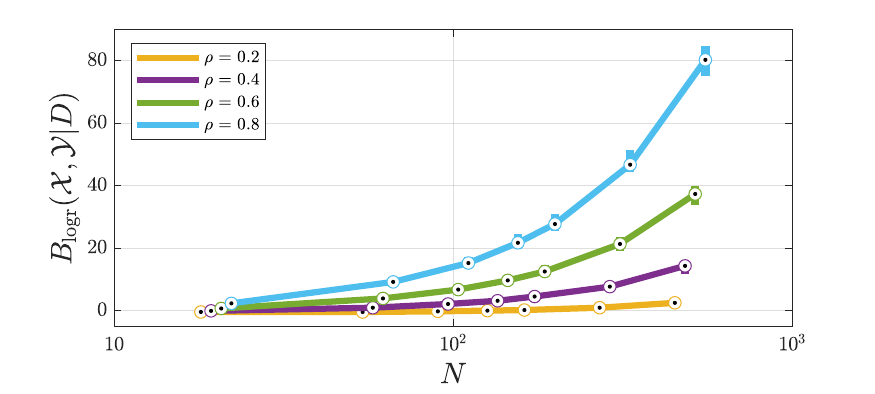} \\
  \vspace{-0.33cm} & \vspace{-0.33cm} \\
  \includegraphics[width=\columnwidth]{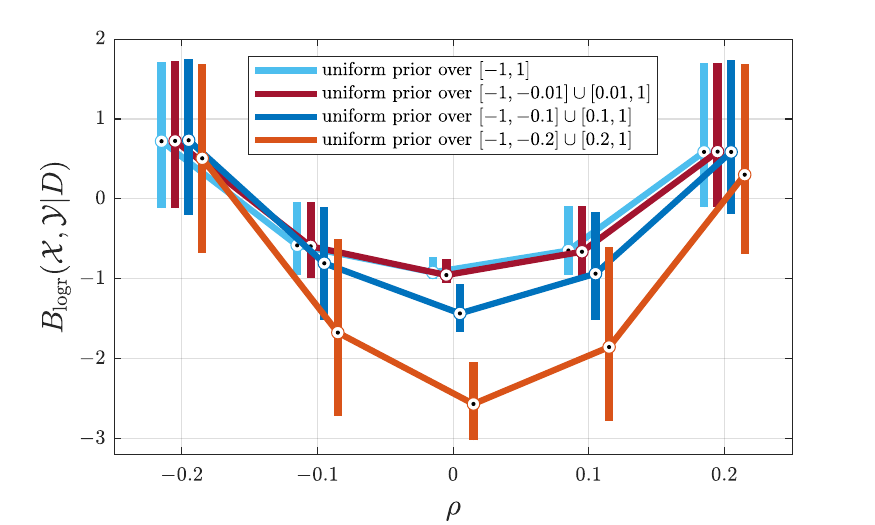}
  & \includegraphics[width=\columnwidth]{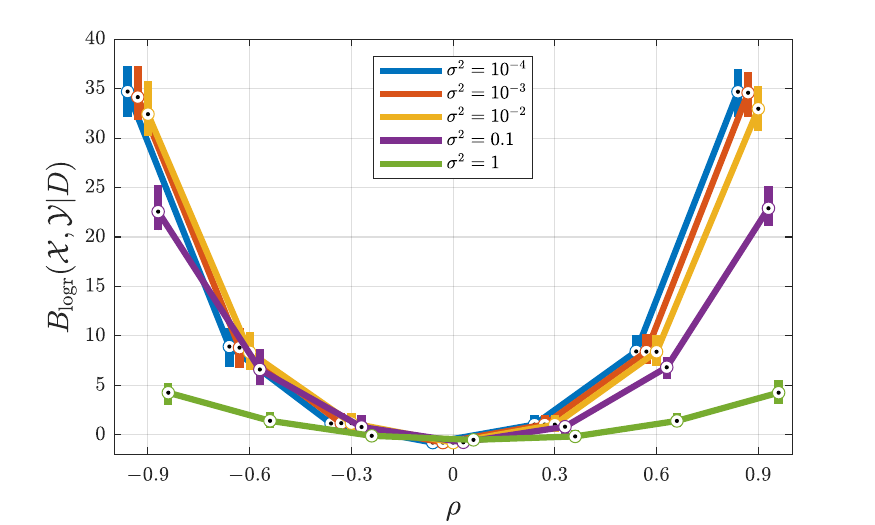}
 \end{tabular}
 \caption{\textbf{Simulation study: bivariate normal distribution with noise.} Boxplots (median and $[ 25\%,75\%]$ percentile) of $\depXYD{logr}$ in various conditions. Top left: Effect of $N$ and $\sigma^2$ for simulations with $\rho =  0$. Top right: Effect of $N$ and $\rho > 0$ for simulations with $\sigma^2 = 0.1$. Bottom left: Effect of prior $\pr ( \rho | H_1 )$ for $\rho \in \{ -0.2, -0.1, 0, 0.1, 0.2 \}$, $\sigma^2 = 10^{-4}$ and $N = 200$. Bottom right: Effect of $\rho$ and $\sigma^2$ for datasets of size $N = 100$.} \label{fig:simu:mbb:res}
\end{figure*}

\subsubsection{Model}

We considered data generated according to the following model: $( X, Y )$ is a bivariate normal distribution with zero mean and covariance matrix given by
\begin{equation}
 \tau ^ 2 \begin{pmatrix}
 1 & \rho \\
 \rho & 1
\end{pmatrix} = \tau ^ 2 M ( \rho ),
\end{equation}
where $\tau$ is assumed to be a known parameter. However, we only measured noisy versions $( U, V )$ of $( X, Y )$ related through
\begin{equation}
  ( U, V ) = ( X, Y ) + ( E, F ),
\end{equation}
with $E$ and $F$ independent Gaussian variables with zero mean and known variance $\sigma ^ 2$. We observed $N$ realizations $( u_n )_{n = 1,\dots, N}$ and $( v_n )_{n = 1,\dots, N}$ of $U$ and $V$. To quantify the dependence between $X$ and $Y$ from these $N$ realizations, we used the following two models: $H_0$, where $X$ and $Y$ are uncorrelated ($\rho = 0$), and $H_1$ where they may be correlated.
\par
The model is analyzed in \supplmat{\S8}, yielding for $\pr ( D | H_0 )$
\begin{equation}
 ( 2 \pi ) ^ { - \frac{ 2 N } { 2 } } \left( \sigma ^ 2 + \tau ^ 2 \right) ^ { - \frac{ 2 N } { 2 } } \exp \left[ - \frac{ \sum_{ n = 1 } ^ N ( u_n ^ 2 + v_n ^ 2 ) } { 2 ( \sigma ^ 2 + \tau ^ 2 ) } \right]
\end{equation}
and for $\pr ( D | H_1 )$
\begin{eqnarray}
  & & ( 2 \pi ) ^ { - \frac{ 2 N } { 2 } } \int \left| \sigma ^ 2 \matr{I} + \tau ^ 2 \matr{M} ( \rho ) \right| ^ { - \frac{ N } { 2 } } \pr ( \rho | H_1 ) \, \ud \rho \\
 & & \times \exp \left[ - \frac{ 1 } { 2 \sigma ^ 2 } \tr \left( \matr{S}  \left\{ \matr{I} - \left[ \matr{I} + \frac{ \sigma ^ 2 } { \tau ^ 2 } \matr{M} ( \rho ) ^ { - 1 } \right] ^ { - 1 } \right\} \right) \right]. \nonumber
\end{eqnarray}
For  $\pr ( \rho | H_1 )$, we considered a prior that could possibly remove a neighborhood of $\rho = 0$. To this end, we used a general distribution of the form
\begin{equation} \label{eq:simu:mbb:prior}
 q_{\epsilon} ( \rho ) = \left\{ \begin{array}{ccl}
  0 & \mbox{for $| \rho | < \epsilon$} \\
  \frac{ 1 } { 2 ( 1 - \epsilon ) } & \mbox{otherwise.}
 \end{array} \right.
\end{equation}
Such a prior imposes $\rho \not \in ] - \epsilon, \epsilon [$ and is uniform otherwise. For $\epsilon = 0$, this is the usual uniform prior on $[ -1,  1]$.

\subsubsection{Data}

We generated data with $\rho$ ranging from $-0.9$ to $+0.9$ by increment of 0.1, $N$ ranging from 20 to 200 by increment of 20 as well as 300 and 500, and $\sigma^2 \in \{ 10^{-4}, 10^{-3}, 10^{-2}, 10^{-1}, 1 \}$. $\tau ^ 2$ was set to 1. For each particular value of the triplet $( \rho, \sigma^2, N )$, we generated $M = 1000$ samples. For each sample, we computed $\depXYD{logr}$ using priors for $\rho$ of the form given in \eqref{eq:simu:mbb:prior} with $\epsilon \in \{ 0, 0.01, 0.1, 0.2 \}$.

\subsubsection{Results}
 
Results are summarized in Fig.~\ref{fig:simu:mbb:res}. For $\rho = 0$ (corresponding to $H_0$ true), $\depXYD{logr}$ was mostly negative and its value decreased with increasing $N$ and decreasing $\sigma ^ 2$ (Fig.~\ref{fig:simu:mbb:res}, top left). For $\rho \neq 0$ (corresponding to $H_1$ true), it was mostly positive, and its value increased with increasing $N$, decreasing $\sigma  ^  2$, and increasing $| \rho |$ (Fig. \ref{fig:simu:mbb:res}, top and bottom right). Furthermore, $\depXYD{logr}$ was observed to behave similarly to a logarithmic function of $N$ for $\rho = 0$ and to a linear function of $N$ for $\rho \neq 0$. Finally, a prior distribution prohibiting small values of $| \rho |$ tended to give more weight to $H_0$ around $\rho = 0$ (Fig.~\ref{fig:simu:mbb:res}, bottom left). These results are in line with our predictions regarding the expected behavior of $\depXYD{}$ when $N$ increases and $\sigma ^ 2$ decreases (see beginning of Section~\ref{s:ss}).

\subsection{Functional dependence with noise} \label{ss:ss:fd}

\begin{figure}[!htbp]
 \centering
  \includegraphics[width=\columnwidth]{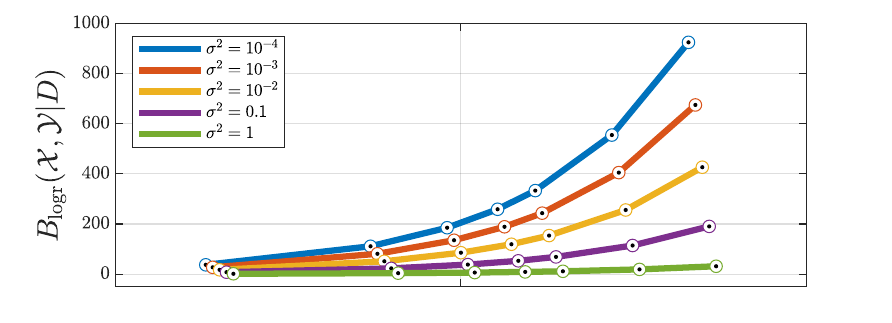} \\
  \vspace{-0.34cm}
  \includegraphics[width=\columnwidth]{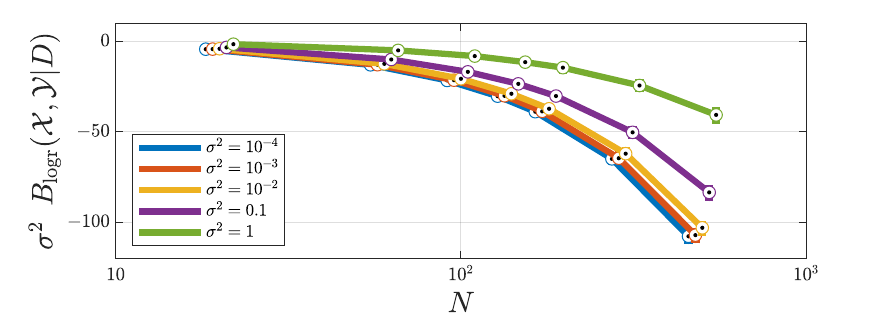}
 \caption{\textbf{Simulation study: functional dependence with noise.} Boxplots (median and $[ 25\%,75\%]$ percentile) of the effect of $\sigma^2$ and $N$ on $\depXYD{logr}$ when the true model is either $H_1$ (top) or $H_0$ (bottom).} \label{fig:simu:rlb:res}
\end{figure}

\subsubsection{Model}

We considered a two-dimensional variable $( X, Y )$ where $X$ and $Y$ may be related by a functional relationship. For the sake of simplicity, we considered a linear relationship. More precisely, for dependence ($H_1$), we assumed that we had
\begin{equation}
  ( X, Y ) = ( T , T ) + ( E, F),
\end{equation}
with $T \sim \mathcal{N} ( 0, \tau ^ 2 )$. By contrast, for independence ($H_0$), we assumed
\begin{equation}
  ( X, Y ) = ( U, V) + ( E, F ),
\end{equation}
with $U, V \sim \mathcal{N} ( 0, \tau ^ 2 )$. In both cases, $E$ and $F$ are white noise of known variance $\sigma^2$, and $\tau ^ 2$ is assumed to be known. Assume that we observed $N$ realizations $( x_n, y_n )_{n = 1,\dots, N}$ of $( X, Y )$. Note that the intermediary variables ($U$ and $V$ for $H_0$; $T$ for $H_1$) are not observed. Interestingly, unlike most models of dependence, the description of dependence here requires fewer parameters (the $t_n$'s) than description of independence (the $u_n$'s and $v_n$'s).
\par
We obtain (see \supplmat{\S9})
\begin{eqnarray}
 & & \ln \frac{ \pr ( D | H_1 ) } { \pr ( D | H_0 )  } \\
 & = & - \frac{ N } { 2 } \ln \left( \sigma ^ 2 \right) - \frac{ N } { 2 } \ln \left( \sigma ^ 2 + 2 \tau ^ 2 \right) + N \, \ln \left( \sigma ^ 2 + \tau ^ 2 \right) \nonumber \\
 & & - \sum_{ n = 1 } ^ N \left[ \frac{ ( x_n - y_n ) ^ 2 } { 2 \sigma ^ 2 ( 2 + \alpha ^ 2 ) } + \frac{ x_n ^ 2 + y_n ^ 2 } { 2 \tau ^ 2 ( 2 + \alpha ^ 2 ) } -  \frac{ x_n ^ 2 + y_n ^ 2 } { 2 \tau ^ 2 ( 1 + \alpha ^ 2 ) } \right], \nonumber
\end{eqnarray}
with $\alpha ^ 2 = \sigma ^ 2 / \tau ^ 2$.

\subsubsection{Data}

We generated data with either model $H_0$ or model $H_1$, $N$ ranging from 20 to 200 by increment of 20 as well as 300 and 500, and $\sigma^2 \in \{ 10^{-4}, 10^{-3}, 10^{-2}, 10^{-1}, 1 \}$. $\tau ^ 2$ was set to 1. For each particular value of $( H_i, \sigma^2, N )$, we generated $M = 1000$ samples. For each sample, we computed $\depXYD{logr}$.

\subsubsection{Results}

Results are summarized in Figure~\ref{fig:simu:rlb:res}. When $H_0$ was true, $\depXYD{logr}$ was found to be negative, a decreasing function of $N$, and an increasing function of $\sigma ^ 2$. When $H_1$ was true, $\depXYD{logr}$ was found to be positive, an increasing function of $N$, and a decreasing function of $\sigma  ^ 2$. Unlike what was found previsouly, $\depXYD{logr}$ behaved similarly to a linear function of $N$ under both $H_0$ and $H_1$. But these results are again in line with our predictions.

\subsection{Dependence through copula} \label{ss:ss:dtc}

\begin{figure}[!htbp]
 \centering
  \includegraphics[width=\columnwidth]{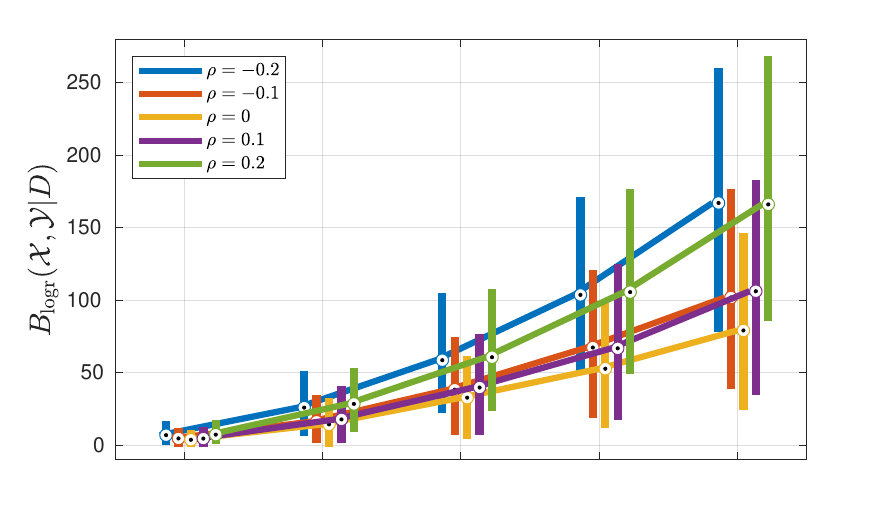} \\
  \vspace{-0.55cm}
  \includegraphics[width=\columnwidth]{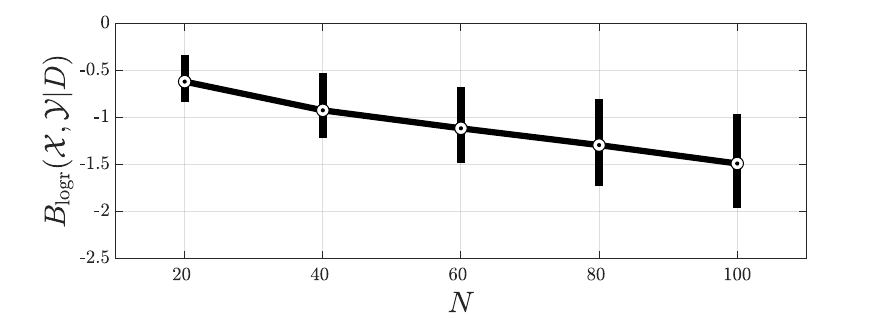}
 \caption{\textbf{Simulation study: dependence through copula.} Boxplots (median and $[ 25\%,75\%]$ percentile) of the effect of $\rho$ and $N$ on $\depXYD{logr}$ when $H_1$ is true (top), and of the effect of $N$ when $H_0$ is true (bottom).} \label{fig:simu:st:dtc}
\end{figure}

\subsubsection{Model}

We modeled a two-dimensional variable $( X, Y )$ with marginals equal to gamma distributions, with $( \alpha, \beta)$ equal to $( 4, 4)$ for $X$ and $( 10, 5 )$ for $Y$. In $H_1$, dependence was modeled through a Student's $t$ copula \gcite[\S8.3]{Ruppert-2015} with 5 degrees of freedom and $\rho \in \{0, 0.2, 0.7 \}$. Note that $\rho = 0$ corresponds to uncorrelated, yet dependent variables \gcite[\S1.16]{Kotz-2004}.

\subsubsection{Data}

We generated data with either model $H_0$ or model $H_1$, $N$ ranging from 20 to 100 by increment of 20. Each time, we generated $M = 1000$ samples. For each sample, we computed $\depXYD{logr}$ using IFD with \eqref{eq:bmod:cm:dep} and \eqref{eq:bmod:cm:im}.

\subsubsection{Results}

Results are summarized in Figure~\ref{fig:simu:st:dtc}. As expected, $\depXYD{logr}$ was found to be (i) negative and a decreasing function of $N$ when $H_0$ was true, and (ii) positive and an increasing function of both $| \rho |$ and $N$ when $H_1$ was true.

\subsection{Dependence of two chaotic systems} \label{ss:ss:dotcs}

\begin{figure}[!htbp]
 \centering
 \begin{tabular}{c}
  \includegraphics[width=\columnwidth]{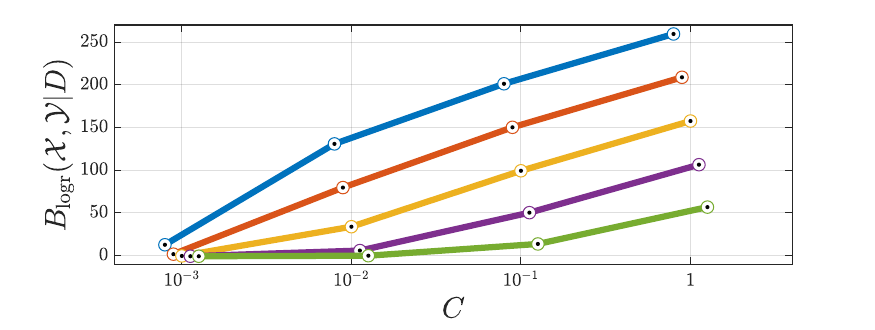} \\
  \vspace{-0.65cm} \\
  \includegraphics[width=\columnwidth]{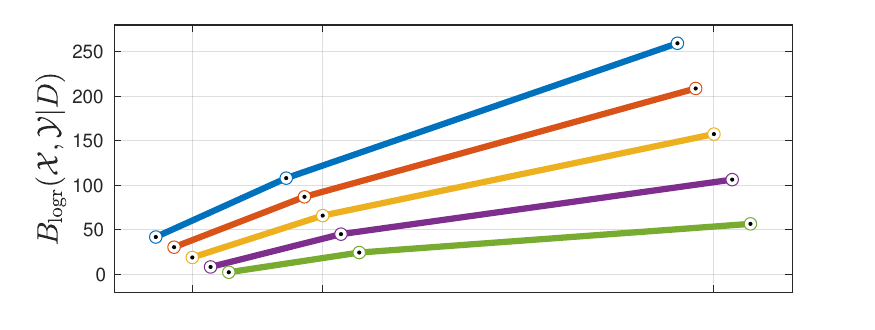} \\
  \vspace{-0.8cm} \\
  \includegraphics[width=\columnwidth]{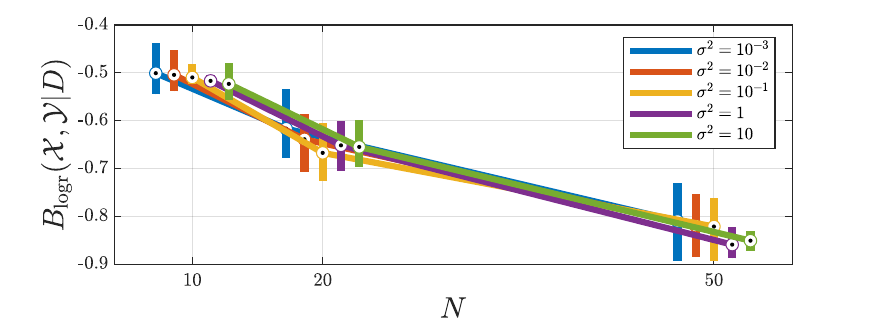}
 \end{tabular}
 \caption{\textbf{Simulation study: dependence of two chaotic systems.} Boxplots (median and $[ 25\%,75\%]$ percentile) of the effect of the coupling parameter $C$ when $N= 50$ (top), and of the effect of $N$ with either $C = 1$, corresponding to $H_1$ true (middle) or with $C = 0$, corresponding to $H_0$ true (bottom).} \label{fig:simu:dotcs}
\end{figure}

\subsubsection{Model}

To demonstrate the possibility of our measure to quantify the intensity of coupling between two systems, we used the example of two coupled chaotic Rössler oscillators with a small parameter mismatch. Each oscillator $i \in \{ 1, 2 \}$ was characterized by its position $(x_i, y_i, z_i)$ and time derivatives  $(\dot{x}_i, \dot{y}_i, \dot{z}_i)$. Coupling was quantified through $C$ ($C = 0$ corresponds to no coupling). For more details, see \supplmat{\S10} or \gcite[\S3.1.4]{Boccaletti-2002}.

\subsubsection{Data}

We simulated data with $C \in \{ 0, 10^{-3}, 10^{-2}, 10^{-1}, 1 \}$. For a given set of parameter values, the trajectory of the system was generated numerically with an explicit Runge-Kutta method and downsampled to one sample per second. Trajectories with $N \in \{ 10, 20, 50 \}$ time points were considered. From each trajectory, $M = 100$ samples were generated by adding Gaussian white noise with variance $\sigma ^ 2 \in \{ 10^{-3}, 10^{-2}, 10^{-1}, 1 \}$.

\subsubsection{Results}

Results are summarized in Figure~\ref{fig:simu:dotcs}. As expected, $\depXYD{logr}$ was globally found to be (i) a decreasing function of $N$ and an increasing function of $\sigma ^ 2$ when $H_0$ was true, and (ii) an increasing function of both $N$ and $C$ and a decreasing function of $\sigma ^ 2$ when $H_1$ was true. Exceptions to this general trend was the case when $C$ was low and $\sigma ^ 2$ large, in which case $\depXYD{logr}$ could first decrease and then increase for increasing $N$.

\subsection{Summary of results} \label{ss:sor}

In the simulation section, we showed that $\depXYD{logr}$ provided a measure of dependence between $X$ and $Y$ that had the following properties:
\begin{itemize}
 \item When $H_0$ was true, $\depXYD{logr}$ typically decreased when the quantity of information available in the data increased (increasing $N$, decreasing $\sigma ^ 2$);
 \item When $H_1$ was true, $\depXYD{logr}$ typically increased when the quantity of information available in the data increased (increasing $N$, decreasing $\sigma ^ 2$). Furthermore, when the intensity of dependence between $X$ and $Y$ was parameterized, $\depXYD{logr}$ was found to be an increasing function of this intensity.
\end{itemize}

\section{Real-life application} \label{s:rla}

Electroencephalography (EEG) is a brain exploration technique that allows to noninvasively record electrical consequences of brain activity. Such recordings are often driven by brain oscillations originating from  synchronized neuronal activity. A common procedure for EEG acquisitions is the so-called event-related protocol, where one records how the brain responds (through the evoked response) to a given stimulation over many repetitions, called trials. For some types of protocols, the stimulus may consistently induce synchronization of brain activity, which translates into a phenomenon called phase resetting. In this case, the phase of the signal (quantified, e.g., through time-frequency analysis) in a certain time window after the stimulus remains consistent over trials. It $\theta_n$ is the phase quantified for trial $n$, $n = 1, \dots, N$, phase consistency has typically been quantified using inter-trial phase coherence (ITC) \gcite{Makeig-2004, Benhamou-2023}, which, in circular statistics, is the mean resultant length $\overline{R}$ of the sample $( \theta_n )$ \gcite[\S2.3.1]{Mardia-2000}
$$\overline{R} = \left| \frac{1}{N} \sum_{ n= 1 }^ N e ^ { i \theta_n } \right|.$$
\par
In our framework, we can propose an alternative measure of the dependence between the stimulus and the brain. More specifically, we conside two competing models $H_0$ and $H_1$, where $H_0$ assume that $\theta_n$ is uniformly distributed on the circle, while $H_1$ assumes that $\theta_n$ has a von Mises distribution with mean direction $\mu$ and concentration parameter $\kappa$ \gcite[\S3.5.4]{Mardia-2000}. Using standard prior distributions for $\mu$ and $\kappa$, we obtain the following measure of dependence (see \supplmat{\S11})
$$\depXYD{logr} = \log_{10} \frac{ \pr ( H_1 ) } { \pr ( H_0 ) } + \log_{10} \left[ \int \frac{ \kappa I_0 ( N \overline{R} \kappa ) }{ \left( 1 + \kappa ^ 2 \right) ^ { \frac{ 3 }{ 2 }} I_0 ( \kappa ) ^ N } \, \ud \kappa \right].$$
In the following, we assume $\pr ( H_1 ) = \pr ( H_0 ) = 1/2$. The integral can be computed numerically for any value of $N$ and $\overline{R}$.

\begin{figure}[!htbp]
 \centering
 \includegraphics[width=0.9\columnwidth]{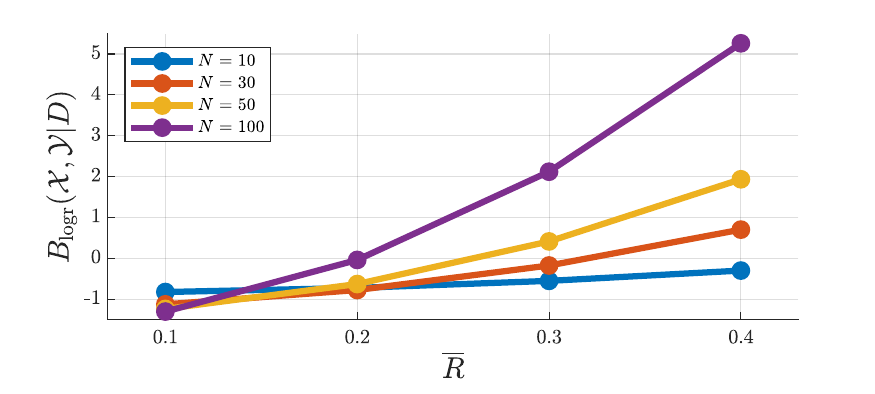} \\
 \includegraphics[width=0.9\columnwidth]{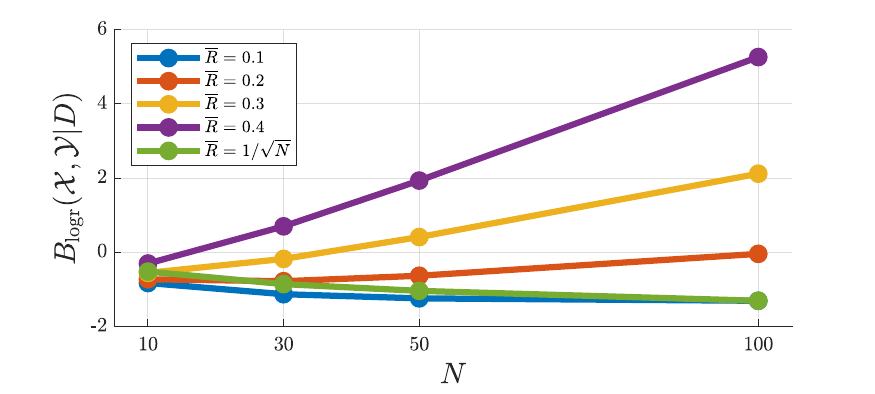} \\
 \includegraphics[width=0.9\columnwidth]{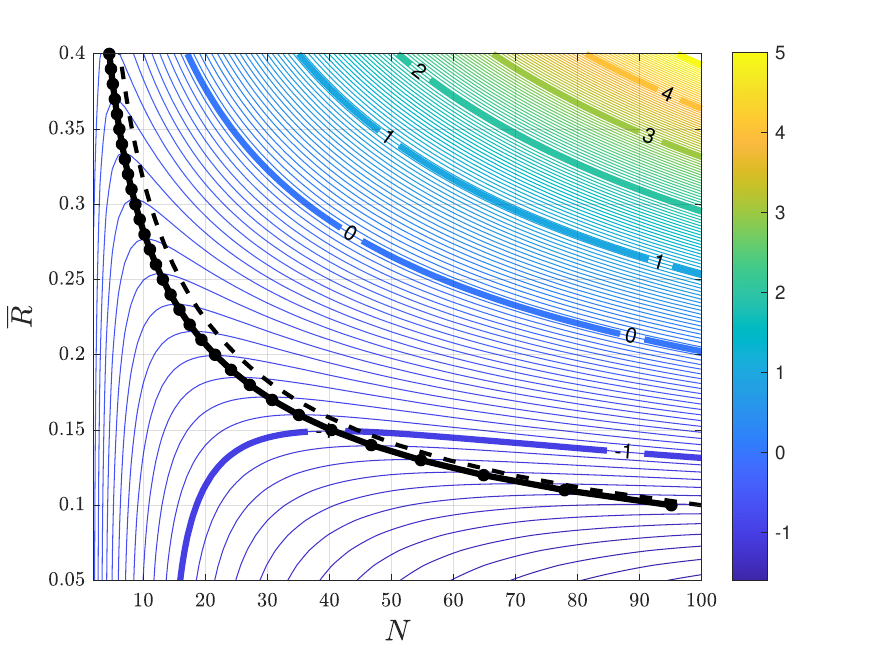}
 \caption{\textbf{Real-life application.} $\depXYD{logr}$ as a function of $\overline{R}$ for various values of $N$ (top) and as a function of $N$ for various values of $\overline{R}$ (midlle). Bottom panel: contour plot of $\depXYD{logr}$, together with $[ N_0 ( \overline{R} ), \overline{R} ]$ (black solid line and circles) and $( N, 1/\sqrt{N} )$ (black dashed line).} \label{fig:ex}
\end{figure}

Results are summarized in Figure~\ref{fig:ex}. $\depXYD{logr}$ was found to be an increasing function of $\overline{R}$, a decreasing function of $N$ for lower values of $\overline{R}$ ($\overline{R} \lessapprox 0.1$), and an increasing function of $N$ for larger values of $\overline{R}$ ($\overline{R} \gtrapprox 0.3$). For intermediate values of $\overline{R}$, $\depXYD{}$ first decreased, then increased. To further investigate this change in monotonicity, we computed for different values of $\overline{R}$
$$N_0 ( \overline{R} ) = \mathrm{argmin}_N \depXYD{logr}.$$
For fixed $\overline{R}$, $\depXYD{logr}$ decreased for $N \leq N_0 ( \overline{R} )$ and increased for $N \geq N_0 ( \overline{R} )$. Since $2 N \overline{R} ^ 2$ is approximately $\chi_2^2$ for large $N$ under the assumption of uniform phase (i.e., $H_0$) \gcite[\S4.4]{Mardia-2000}, we have $\esp ( \overline{R}^2 | H_0 ) \approx 1 / N$, that is, we can expect $\overline{R}$ to have values of the order of $1 / \sqrt{N}$ (still under $H_0$). We empirically noticed that $N_0 ( \overline{R} )$ was smaller than, yet close to $1 / \sqrt{N}$.

\section{General properties of $\depXYD{}$} \label{s:gp}

Various researchers have tried to define a set of properties that a good measure of dependence should respect \gcite{Goodman_LA-1954, Renyi-1959b, Hall-1967, Jupp-1980, Maasoumi-2002, Granger-2004, Dionisio-2006}. The properties of $\depXYD{}$ exhibited so far either in the calculations (Section~\ref{s:bmod}) or the stimulation study (Section~\ref{s:ss}) are closely related to these properties, namely:
\begin{enumerate}
 \item It is well defined for any pair of (either continuous or discrete) variables $X$ and $Y$ as long as the corresponding models $H_0$ and $H_1$ are;
 \item It is symmetrical in $X$ and $Y$ as long as both $H_0$ and $H_1$ share this property;
 \item It reaches its minimum value when $X$ and $Y$ are known to be independent as described by $H_0$;
 \item It reaches its maximum value when $X$ and $Y$ are known to be dependent as described by $H_1$;
 \item It is an increasing function of mutual information, when such a measure makes sense;
 \item It is an increasing function of the linear correlation coefficient in the case where $( X, Y )$ follows a bivariate normal distribution.
\end{enumerate}
Properties 3 and 4 are quite restrictive, since we assume that we \emph{know} what the underlying model is. Indeed, we usually do not know for sure whether $H_0$ or $H_1$ is true. In this case, we must set $\Pr ( H_0 ) \neq 0$ and $\Pr ( H_1 ) \neq 0$. Results from both the theoretical calculations and the simulation studies indicate that the following original properties also hold asymptotically:
\begin{enumerate}
 \item[3')] If $H_0$ is true, then $\depXYD{}$ is a decreasing function of $N$ which tends to its lower bound as $N \to \infty$;
 \item[4')] If $H_1$ is true, then $\depXYD{}$ is an increasing function of $N$ which tends to its upper bound as $N \to \infty$;
\end{enumerate}
\begin{enumerate}
 \item[7)] In case of model misspecification, $\depXYD{}$ behaves as if the true model were the one closer to the true generative model (in the sense of Kullback--Leibler divergence).
 \end{enumerate}
 Note that, from all the results presented in this manuscript, Property~3' seemed to hold even for small values of $N$, while Property~4' sometimes required larger values of $N$ to hold and followed an initial stage where $\depXYD{}$ decreased. Finally, we observed the following empirical properties from the simulation studies:
 \begin{enumerate}
  \item[8)] The effect of noise was the reverse of the effect of $N$: it tended to increase $\depXYD{}$ when $H_0$ was true and to decrease it when $H_1$ was true.
 \item[9)] When $H_1$ was true and the intensity of dependence was quantified by a parameter, $\depXYD{}$ was an increasing function of this intensity.
\end{enumerate}
To our knowledge, it is the first time that these properties are considered as potentially desirable features for a measure of dependence. This is further discussed in the next section.

\section{Discussion} \label{s:disc}

\paragraph*{Summary}

Quantification of dependence between two systems is still an open issue in the general case. We here proposed a general measure of \modif{dependence} $\depXYD{}$ between two systems $\syst{X}$ and $\syst{Y}$ for a given dataset $D$ based on the Bayesian comparison of two models, one of independence $H_0$ and another one of dependence $H_1$. \modif{Dependence} within $D$ was then quantified as the posterior probability of $H_1$ given $D$, $\depXYD{pr} = \Pr ( H_1 | D )$, or a strictly increasing function of it. We calculated the value of $\depXYD{lnr} = \ln \depXYD{pr}$ in particular cases: when the model distribution is known, when it belongs to a known parametric family with unknown parameters, when $H_0$ is nested in $H_1$ (including common models such as maximum-entropy distributions, multivariate normal distributions, and bivariate discrete distributions), and when  dependence is modeled through a copula. Using simulations, we investigated the behavior of $\depXYD{logr} = \log_{10} \depXYD{pr}$ in the presence of noise and when $H_1$ had a parameter coding for the intensity of dependence. We also provided an application of our framework to neuroscience and neuroimaging. Finally, we stated some key general properties of $\depXYD{}$. While some of these properties are typical of what is expected from a valid measure of dependence, others are novel and naturally appear as desirable features for $\depXYD{}$.

\paragraph*{Existing work}

The theoretical results introduced here relating the posterior distribution and its variants to mutual information (Section~\ref{s:bmod}) bring together two existing lines of research from standard statistical theory and information theory. On the one hand, the behavior of the model marginal likelihood for large $N$ has been studied in depth and the theoretical underpinnings of such calculations go back to the development of the BIC and the minimum description length (MDL). On the other hand, the asymptotic bias of empirical mutual information estimators is also well known. The connection between Bayesian posterior distribution and mutual information was first noticed in particular cases by \gcitet{Wolf_DR-1994}, \gcitet{Marrelec-2015} and \gcitet{Marrelec-2021b}. We here showed that this connection actually holds in the more general setting of nested models (Section~\ref{ss:bmod:nm}). While these results involve developments that are rather standard in statistical theory and information theory, they shed some important light on the relevance of $\depXYD{}$ as a valid measure of \modif{dependence}.
\par
To our knowledge, no general theory exists regarding Bayesian model comparison for non-nested models. Our results from copula modeling of dependence (Sections \ref{ss:bmod:cm} and \ref{ss:ss:dtc}) suggested that similar results regarding the behavior of $\depXYD{}$ might hold in that particular case.

\paragraph*{Difference of behavior under $H_0$ and $H_1$}

It has to be underlined that the roles of $H_0$ (independence) and $H_1$ (dependence) are not symmetrical. Often, $H_0$ is nested in $H_1$ or, equivalently, $H_1$ includes $H_0$ as a particular case (parameters set to particular values, e.g., usually 0). This has the following consequence for Bayesian model comparison. In the case of independence, a $H_1$ with likely parameter values becoming increasingly closer to, e.g., 0 cannot be fully ruled out. By contrast, in the case of dependence, $H_0$ becomes increasingly unlikely as $N$ increases. This translates into a typically different behavior of $\depXYD{}$: a logarithmic decrease under $H_0$ but a linear increase under $H_1$. Such a behavior, which was observed in all our computations as well as our first example in the simulation study, was associated with a connection with mutual information. By contrast, the second example of our simulation study (Section~\ref{ss:ss:fd}) exhibited linear behaviors both when $H_0$ and $H_1$ were true, together with a connection with mutual information that remains to be determined.
\par
Another difference between $H_0$ and $H_1$ can be seen in the monotonic property of $\depXYD{}$ with respect to $N$. As detailed in various places in the manuscript, $\depXYD{}$ under $H_0$ was found to be a decreasing function of $N$ even for moderate to small values of $N$. By contrast, under $H_1$, it cannot be ruled out that $\depXYD{}$ has an initial stage where it is actually a decreasing function of $N$, before becoming an increasing function. We interpret it as a consequence of the fact that Bayesian analysis tries to find a compromise between the strength of dependence and model parsimony, in line with the usual interpretation of Bayesian inference as providing a quantitative implementation of Occam's razor. How this non-monotonic behavior fits in our general framework remains to be further clarified.

\paragraph*{Measures of dependence}

The main point of the present work is that $\depXYD{}$ is a valid measure of dependence. Quantification of dependence is still a field of ongoing research, whose objective is to provide a measure that quantifies the departure of $f_{XY} ( x, y )$ from independence, i.e., from $f_X ( x ) f_Y ( y )$. To our knowledge, the present work is the first one to advocate that the posterior probability in a specifically designed Bayesian model comparison analysis can be considered as a valid measure of dependence. Importantly, $\depXYD{}$, as a result of a Bayesian analysis, quantifies by construction \modif{the evidence for} dependence between $\syst{X}$ and $\syst{Y}$ as modeled by $H_1$ (compared to $H_0$) and observed in $D$. It is what we would call an \emph{inferential} measure of dependence, in that it is both \emph{model-based}---as it incorporates information from (probabilistic) models $H_0$ and $H_1$---and \emph{data-driven}---as its value reflects the content of a dataset $D$. \modif{As a consequence, the same value of $\depXYD{}$ can be obtained for two very different scenarios: either a large amount of data about weakly dependent variables, or a small amount of data about strongly associated variables. Yet, for a given system, changes in  $\depXYD{}$ can be interpreted unambiguously: W}hen the information content of $D$ increases (e.g., with increasing size  or decreasing noise), the measure becomes increasingly closer to the boundary of its definition domain that corresponds to the correct ideal situations, in agreement with Properties 3, 3', 4, and 4' of Section~\ref{s:gp}. Importantly, this is not a direct consequence of the fact that we used a Bayesian model comparison analysis but is the result of a selective choice of the measure. For instance, a non-monotonic function of $\Pr ( H_1 | D )$ such as $\frac{1}{N} \ln \Pr ( H_1 | D )$, which behaves in a fashion very similar to mutual information, would not qualify as an inferential measure of dependence (see \supplmat{\S12}). We will come back to this point below.
\par
In parallel to the development of new measures of dependence, there has also been active research regarding the properties that a good measure of dependence should respect \gcite{Goodman_LA-1954, Renyi-1959b, Hall-1967, Jupp-1980, Maasoumi-2002, Granger-2004, Dionisio-2006}. In Section~\ref{s:gp}, we provided properties that hold for $\depXYD{}$. Properties 1--6 are based on previous descriptions of how ideal measures of dependence should behave, while Properties 3', 4', 7, 8, and 9 have been introduced in this manuscript and are specific to the expected behavior of what we coined inferential measures of dependence.
\par
Some authors also consider that a good measure of dependence should be a metric \gcite{Maasoumi-2002, Granger-2004, Dionisio-2006}. We did not check this requirement, and it would be of interest to see whether it can be met by $\depXYD{}$. Still, according to the main properties of a dependence measure, the more two variables are dependent, the larger the measure. In classification terminology, this makes a dependence measure closer to a similarity measure than to a distance measure. To our knowledge, this is in agreement with only one instance of measure proposed in the literature \gcite{Kraskov-2005}, which is indeed a decreasing function of dependence and, therefore, does not meet Properties 3--6.

\paragraph*{Posterior probability and mutual information}

In this manuscript, we made several connections between $\depXYD{}$ and the plug-in estimator of mutual information $\hat{I}(X,Y)$, showing that $\depXYD{}$ is often, even approximately, an affine function of $\hat{I}$. Still, there are some major differences between $\depXYD{}$ and $\hat{I}(X,Y)$---and, more generally, mutual information---that need to be stated and clarified.
\par
First, we argue that $\depXYD{}$ is more general than mutual information in two senses. First, while there are many cases where there is a (direct or indirect) connection between $\depXYD{}$ and mutual information, there are also cases where such a connection may not make sense or does not exist (see, e.g., the simulation study, Section~\ref{s:ss} and \supplmat{\S13}). In such cases, we advocate that the use of $\depXYD{}$---instead of mutual information---is justified by the framework we introduced in the present manuscript and still makes sense. Second, unlike mutual information, whose value taken by an estimator on a given dataset of finite size does not have any meaning, the values taken by Bayesian measures of dependence  have a simple interpretation, as they quantify the evidence in favor of dependence for the dataset under consideration.
\par
Another major difference between $\depXYD{}$ and $\hat{I} ( X, Y )$ is that they are measures of dependence of different nature. $\depXYD{}$ is obtained through hypothesis testing and quantifies the evidence of $H_1$ against $H_0$. By contrast, mutual information quantifies the theoretical level of dependence contained in a model $H_1$ compared to $H_0$. It is usually a function of model parameters that need to be estimated to provide an estimate of mutual information. As a consequence, when $H_1$ is true (dependence), (i) $\esp [ \hat{I}(X,Y) ]$ is not necessarily an increasing function of $N$ as $N \to \infty$, and (ii) it does not tend to the upper bound of its range ($+ \infty$) as $N \to \infty$. More precisely, in the case of independence, $H_0$ can often be associated with a zero mutual information (Sections \ref{ss:bmod:kd}--\ref{ss:bmod:nm}). In this case, any valid estimator of mutual information will tend to 0 as $N \to \infty$. Since 0 is the lower bound of mutual information, Properties 3 and 3' can be expected to roughly hold. By contrast, if $H_1$ is true (dependence), any valid estimator of mutual information will tend to the theoretical value of mutual information (which is in general strictly lower than its upper bound, $+ \infty$), and changes in the estimator values will be mainly due to statistical fluctuations around this theoretical value. As a consequence, Properties 4 and 4' of Section~\ref{s:gp} are not respected.
\par
A model-centered measure aims at quantifying the theoretical level of dependence between two variables entailed by, or contained in, a model (and not data). It ranks models, from one(s) with the least dependence (usually independence, for which the measure reaches its lower bound) to one(s) with the most dependence (for which the measure reaches its upper bound). Data are then used to infer this theoretical level of dependence \emph{from} them. Such a feature is not specific to mutual information, and the family of model-centered measures include many other existing measures of dependence, including all those mentioned in the introduction (maximal correlation coefficient \gcite{Gebelein-1941, Renyi-1959b}, the mixed derivative measure of marginal interaction \gcite[\S2.3]{Whittaker-1990}, Hoeffding's procedure \gcite{Hoeffding-1948}, distance correlation \gcite{Szekely-2007}, circular correlation \gcite{Jupp-1980}, Hilbert--Schmidt information criterion \gcite{Gretton-2005b})

\paragraph*{Limitations of method}

The current approach strongly relies on Bayesian model comparison. As a consequence, it has limitations that are typical of that kind of approaches and are related to the choice of the model, the choice of the priors, and the computation of the marginal model likelihoods.
\par
First, our approach requires the specification of two models, one for independence ($H_0$) and one for dependence ($H_1$). Of course, it cannot be ruled out that either, or both models are incorrect, hence the importance of considering model misspecification. In that case, it is possible to show (see Section~\ref{ss:bmod:mm}) that, under specific assumptions, $\depXYD{}$ will behave as if the true model were the one that is closer (in the sense of Kullback--Leibler divergence) to the true generative one. This is in line with existing general results regarding Bayesian model comparison \gcite[\S7.27]{OHagan-2004}. Note that the choice of $H_0$ is often dictated by the type of data considered. It is somewhat made easier by the (strong) constrain of \eqref{eq:indep:def}. By contrast, the choice for $H_1$ should take into account both the type of data and the potential structure of dependence. As a consequence, we expect this choice to be more complex and prone to successive adjustments.
\par
Also, Bayesian analyses require the introduction of prior distributions. Here, we needed prior information regarding both the relative plausibility of $H_0$ and $H_1$  as well as the potential values of the parameters for both models. As data size grows, we expect the respective priors for $H_0$ and $H_1$ to have vanishing impact on $\depXYD{}$, unless there is prior evidence that one model is overwhelmingly more plausible than the other. By contrast, choosing the priors on the model parameters is more problematic. Noninformative priors, which are commonly used in Bayesian parameter estimation for the sake of simplicity, are strongly advised against for model comparison. Conjugate priors, which are often used, might not correctly represent the prior information at hand, while using tailored priors might lead to an intractable $\depXYD{}$. See \gcitet{Marrelec-2015} for a discussion of the choice of the prior on the covariance matrix for the multivariate normal model. Unlike Bayesian parameter estimation, where the impact of the prior vanishes for large $N$, Bayesian model comparison is influenced by the choice of the model parameter prior for any data size. Still, the manuscript provides two results that specify the role of this type of prior. First, we theoretically showed that, with increasing data size, it is the dimension of the parameter space that matters rather than the distribution of its values in that parameter space (see Sections~\ref{ss:bmod:klfwup}--\ref{ss:bmod:nm}), in line with the usual Laplace and BIC approximations. Since Bayesian analysis can involve very complex models with many parameters in high dimension---in particular when modeling dependence in $H_1$---, these asymptotic results may be of limited relevance in some applications. In these cases, one needs to go back to the exact formulas with finite $N$. The parameter prior may then have an effect on $\depXYD{}$, an effect that it is important to quantify, e.g., through sensitivity analysis using different prior distributions. For instance, we empirically showed on simulated data [Section~\ref{ss:ss:mbb}, see in particular $q_{\epsilon} ( \rho )$ and \eqref{eq:simu:mbb:prior}] that having a more specific prior had limited influence when $H_1$ was true and the model parameter was indeed in the correct range, while helping ruling out $H_1$ when $H_0$ was true.
\par 
Finally, the broad applicability of $\depXYD{}$ will partly depend on the ease with which ratios of marginal model likelihoods can be computed. Exact calculation of marginal model likelihoods is usually very difficult, from both an analytical and a computational point of view. Analytical calculations often rely on simple models combined with conjugate priors (as in the present manuscript), tricks, such as the Savage--Dickey density ratio \gcite{Dickey-1971}, or approximations, e.g., Laplace approximation \gcite{Tierney-1986, Gelfand-1994} or the  BIC \gcite{Schwarz-1978}. Exact computational approaches are based on numerical integration, which is time-intensive and suffers from the curse of dimensionality. A wealth of approximate methods have been devised, with specific advantages and limits, which make them more or less suitable depending on the context: the acceptance ratio method and thermodynamic integration \gcite{Neal-1993}; importance sampling, bridge sampling and path sampling \gcite[\S5]{Chen_MH-2000}; reversible jump Markov chain Monte Carlo (RJ-MCMC) \gcite{Green-1995}; nested sampling \gcite{Skilling-2004}.

\paragraph*{Generalization to several variables}

In the present manuscript, we focused on the quantification of dependence between two (potentially multidimensional) variables. The framework can very easily be expanded to take into account dependence between several variables. The general framework is the same, and was detailed in \gcitet{Marrelec-2021b}. The connection between Bayesian measures of dependence and mutual information still holds, with mutual information replaced by a generalization to several variables known as total correlation \gcite{Watanabe-1960}, multivariate constraint \gcite{Garner-1962}, $\delta$ \gcite{Joe-1989}, or multiinformation \gcite{Studeny-1998}.

\paragraph*{Questioning the notion of dependence}

Importantly, the use of our framework made it clear that independence itself is not enough to provide an unambiguous measure of dependence. Indeed, there was a need to introduce both an alternative model $H_1$, in which a potential structure of dependence was introduced, and data,  which where used to quantify dependence. Both the model of dependence and the data have a key influence on the resulting dependence measure.
\par
It cannot be excluded that data that appear to be associated to independent variables are actually dependent but with a dependence structure that is not the one introduced in $H_1$. This could appear as a weakness of the method, but we believe it is rather a strength of it: While the underlying assumptions (in particular regarding the model of dependence $H_1$) are implicit in many methods, here they have to be clearly stated and translated into operational form to be able to perform a Bayesian analysis. We argue that this makes the model easy to falsify and, as a consequence, to improve. It is not uncommon that variables that were believed to be independent are later found dependent based on a different model of the dependence pattern. For instance, in neuroimaging, where the authors have extended experience, there has been a major interest in the subfield of functional connectivity analysis, where one tries to use brain imaging data to quantify the \modif{dependence} between brain regions. Unsurprisingly, functional connectivity is quantified differently based on the imaging modality. In functional magnetic resonance imaging (fMRI) \gcite{Huettel-2004b, Buxton-2009}, the data are commonly assumed to follow a normal distribution and dependence is often quantified through pairwise correlation \gcite{Marrelec-2006, Bellec-2012b}. By contrast, in electroencephalography  (EEG) and magnetoenccephalography (MEG) \gcite{Hari-2017}, pairwise correlation is considered as a poor measure of dependence, and other, more adapted measures of functional connectivity have been proposed, based on models of circular or oscillating data \gcite{Marrelec-2005b, van_Diessen-2015}. Another example is the real-life application of Section~\ref{s:rla}), where the two advantages of $\depXYD{}$ appear clearly: (1) the underlying assumptions are made apparent (uniform vs. von Mises phase), and (2) these assumptions can be checked, falsified, and possibly changed for more realistic modeling.
\par
Paralleling what was said for $H_1$, it cannot be excluded that variables that are associated with low or decreasing values of $\depXYD{}$ based on small datasets would actually yield large or increasing  values of $\depXYD{}$ with larger data sets. As mentioned above, Bayesian analysis comes with a built-in quantitative implementation of Occam's razor which performs a compromise between the strength of dependence and model parsimony. As a consequence, data originating from weakly dependent models may at first yield low or decreasing values of $\depXYD{}$. Such a behavior emphasizes the importance of considering $\depXYD{}$ as a measure of \modif{the \emph{evidence for dependence in a given dataset with a given model of dependence}.}

\section{Conclusion}

The present work is the first one to advocate that the posterior probability (or any strictly increasing measure thereof) resulting from a specifically designed Bayesian model comparison analysis can be considered as a valid measure of dependence. We showed that such a framework provided a family of measures, denoted $\depXYD{}$ that quantify the information contained in $D$ in favor of $H_1$ versus $H_0$. As such, they quantify the evidence for---or credibility of---dependence between $\syst{X}$ and $\syst{Y}$ as modeled by $H_1$ (compared to $H_0$) and observed in $D$. All measures in this family shared the following key asymptotic properties in a wide range of situations:
\begin{itemize}
 \item Under $H_0$, $\depXYD{}$ is a decreasing function of $N$ which tends to its lower bound when $N \to \infty$;
 \item Under $H_1$, $\depXYD{}$ is an increasing function of $N$ for $N$ large enough (after a potential initial stage of decrease) and tends to its upper bound when $N \to \infty$;
 \item $\depXYD{}$ is an increasing function of some common existing measures of dependence, such as correlation, mutual information, the minimum discrimination information, and the log-likelihood ratio criterion for testing independence;
 \item In case of model misspecification, $\depXYD{}$ behaves as if the true model were the one closer to the true generative model.
\end{itemize}
Empirically, we also showed that (i) increasing noise had the opposite effect on $\depXYD{}$ to increasing $N$, in that it drew the measure away from the expected (lower for $H_0$, upper for $H_1$) bound, and (ii) when $H_1$ had a parameter coding for the intensity of dependence, $\depXYD{}$ was an increasing function of this intensity. Finally, the value computed for a given dataset of finite size has a direct, exact interpretation as (a strictly increasing function of) the probability that $\syst{X}$ and $\syst{Y}$ be dependent (as described by $H_1$) for that given dataset. Our objective is now to show the generality and versatility of $\depXYD{}$ as \modif{an inferential measure} of dependence.


%




\section*{Acknowledgment}

The authors are grateful to the reviewers for their comments which substantially improved the quality of the manuscript.

\ifCLASSOPTIONcaptionsoff
  \newpage
\fi



\bibliographystyle{IEEEtran}
\bibliography{abrev,anglais,mabiblio}

\vfill


\end{document}